\documentclass[sigconf]{acmart}
\AtBeginDocument{%
  \providecommand\BibTeX{{%
    \normalfont B\kern-0.5em{\scshape i\kern-0.25em b}\kern-0.8em\TeX}}}

\acmISBN{978-1-4503-XXXX-X/18/06}

\usepackage{multirow}
\usepackage{bm}
\usepackage{graphicx}
\usepackage{textcomp}
\usepackage{xcolor}
\usepackage{pifont}
\usepackage{makecell}
\usepackage{algorithm}
\usepackage{algpseudocode}
\usepackage{ulem}
\usepackage[capitalise,noabbrev]{cleveref}
\usepackage{subcaption}
\usepackage{pifont}
\usepackage{gensymb}
\usepackage{makecell}

\usepackage[most]{tcolorbox}

\newcommand{\ourmethod}{ADO-LLM}
\newcommand{\xmark}{\ding{55}}%
\newcommand{\gp}{\mathcal{GP}}
\DeclareMathOperator*{\argmax}{argmax}

\copyrightyear{2024} 
\acmYear{2024} 
\setcopyright{rightsretained} 
\acmConference[ICCAD '24]{IEEE/ACM International Conference on
Computer-Aided Design}{October 27--31, 2024}{New York, NY, USA}
\acmBooktitle{IEEE/ACM International Conference on Computer-Aided Design
(ICCAD '24), October 27--31, 2024, New York, NY, USA}
\acmDOI{10.1145/3676536.3676816}
\acmISBN{979-8-4007-1077-3/24/10}
\begin{document}

\title{ADO-LLM: Analog Design  Bayesian Optimization with In-Context Learning of Large Language Models}
\author{Yuxuan Yin$^\ddag$, Yu Wang$^\ddag$, Boxun Xu, Peng Li$^*$}
\email{{y_yin, yu95, boxunxu, lip}@ucsb.edu}
\affiliation{
  \institution{Department of Electronic and Computer Engineering, University of California, Santa Barbara}
  \state{California}
  \country{USA}
}

\begin{abstract}
Analog circuit design requires substantial human expertise and involvement, which is a significant roadblock to design productivity.  Bayesian Optimization (BO), a popular machine-learning-based optimization strategy, has been leveraged to automate analog design given its applicability across various circuit topologies and technologies. Traditional BO methods employ black-box Gaussian Process surrogate models and optimized labeled data queries to find optimization solutions by trading off between exploration and exploitation. However, the search for the optimal design solution in BO can be expensive from both a computational and data usage point of view, particularly for high-dimensional optimization problems.  This paper presents ADO-LLM, the first work integrating large language models (LLMs) with Bayesian Optimization for analog design optimization. 
ADO-LLM leverages the LLM’s ability to infuse domain knowledge to rapidly generate viable design points to remedy BO's inefficiency in finding high-value design areas specifically under the limited design space coverage of the BO's probabilistic surrogate model. In the meantime, sampling of design points evaluated in the iterative BO process provides quality demonstrations for the LLM  to generate high-quality design points while leveraging infused broad design knowledge. Furthermore, the diversity brought by BO's exploration enriches the contextual understanding of the LLM and allows it to more broadly search in the design space and prevent repetitive and redundant suggestions.
We evaluate the proposed framework on two different types of analog circuits and demonstrate notable improvements in design efficiency and effectiveness. 
\end{abstract}

\maketitle

\def\thefootnote{$\ddag$}\footnotetext{Equal contribution.}

\def\thefootnote{$*$}\footnotetext{Corresponding author.}

\renewcommand{\shortauthors}{Yuxuan Yin, Yu Wang, Boxun Xu, Peng Li}





\section{Introduction}

Analog circuit sizing is a critical yet challenging task in electronic design automation, characterized by
a vast and intricate design space that requires extensive human expertise and involvement. The multi-objective nature of the design process introduces further complications.  Designers need to find a delicate balance among competing objectives such as power efficiency, area minimization, and performance maximization.  Moreover, each shift in circuit topology or technology requires a reevaluation of established design principles, adding to the complexity of achieving optimal designs.

Bayesian Optimization (BO) has emerged as a powerful machine learning tool for tackling analog circuit sizing due to its ability to efficiently navigate large and complex search spaces. BO employs a principled probabilistic surrogate model, typically a Gaussian process (GP), to estimate the performance of various circuit design points and relies on the minimization of a well-defined acquisition function to propose new design points, which balances the exploration and exploitation in the search space. The new design points are then simulated to obtain performance metrics that provide additional labeled training data to refine the surrogate model in an active learning manner. This method has been notably advanced in recent lines of work such as \cite{mace}, \cite{phcbo}, and \cite{demo}.

Despite these advancements, BO is not without its limitations. Firstly, as a black-box optimization method, it intrinsically lacks domain-specific analog design knowledge and merely searches for the best mathematically defined figure-of-merit (FOM). It does so without exploring other informative feedback provided by the circuit simulator such as transistor regions of operation that can shed light on key circuit structural properties underlying optimized FOM and robustness of the circuit.   As such, it does not exploit all available opportunities in the iterative design process for fast design convergence.  
Secondly, BO is well suited for single-objective optimization tasks but can struggle with the multi-objective nature of analog circuit sizing, often failing to capture the nuanced trade-offs between competing design objectives required for optimal solutions. 
While these limitations are discussed within the framework of Bayesian optimization, they are indicative of a broader challenge faced by many machine learning models. Specifically, they highlight the inherent difficulty in effectively integrating a broad range of design knowledge when the models are trained solely on a limited amount of domain-specific data.


Large Language Models (LLMs) present great promise in addressing the above challenge and may be well-positioned to augment BO’s capabilities in analog circuit sizing tasks. We argue that the power of LLMs in circuit design originates from two key capabilities: the extensive prior knowledge embedded in their pre-training data and their ability to perform in-context learning that can be utilized to incorporate domain expertise. This allows LLMs to suggest design modifications and innovations with a high degree of relevance. Typically, the application of LLMs in this context follows an iterative loop, where previous design examples are demonstrated to the LLM, prompting it to generate new, potentially optimal design points. These points are then simulated, and the outcomes are used to further enrich the LLM’s training examples  \cite{ladac}. However, this few-shot generation process has limitations. The quality of optimization heavily depends on both the intrinsic capabilities of the LLM and the quality of the input demonstrations. Moreover, LLMs tend to produce solutions that closely mimic the provided examples, showing a hesitance to extrapolate beyond the demonstrated examples and explore new design areas.

\begin{figure*}[htbp]
    \centering
    \includegraphics[width=1\linewidth]{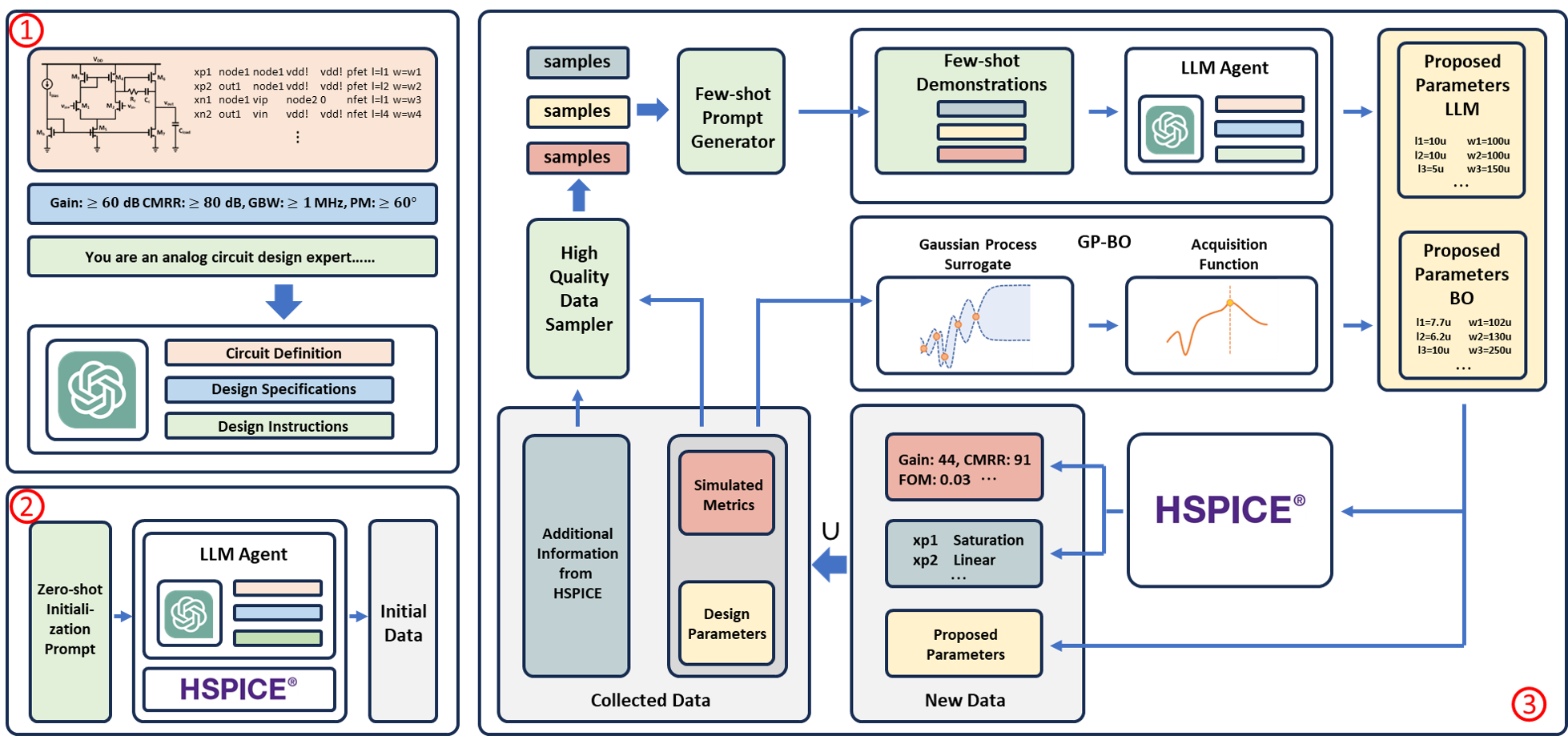}
    \caption{Overview of the ADO-LLM framework. Top-left: the LLM agent is initialized with the circuit definition, the design specifications, and the design instructions. Bottom-left: the LLM leverages the inherent domain knowledge to generate viable initial design points. Right: the optimization flow of ADO-LLM in each iteration.}
    \label{fig:overview}
\end{figure*}

In this paper, we introduce a novel unified framework named ADO-LLM, which combines Large Language Models (LLMs) with Bayesian Optimization (BO) to leverage the strengths of both methods while overcoming their respective limitations. 
ADO-LLM comprises three primary components: a standard Gaussian Process-based Bayesian Optimization (GP-BO) proposer, an LLM agent, and a high-quality data sampler. The GP-BO proposer operates on the entirety of  a collected dataset consisting of all design points evaluated so far, proposing new points for additional evaluation.  In the meantime, the LLM agent focuses on a subset of high-quality demonstrations sampled from this shared dataset to generate new design suggestions. These proposed design points are then evaluated using a circuit simulator and added to the dataset, following the standard iterative process of BO.

Within the ADO-LLM framework, BO benefits from the LLM’s ability in infusing domain knowledge to rapidly generate viable design points that improve the quality of the shared dataset. This remedies BO's inefficiency in finding high-value design areas during exploration, specifically when the scarcity of labeled training data limits the design space coverage of the BO's probabilistic GP surrogate model. With the aid of the LLM, BO continues to explore promising yet under-explored areas using the GP surrogate, proposing diverse new design points to be evaluated. Concurrently, the BO helps provide quality demonstrations sampled from the shared common dataset to the LLM; this quality assurance in demonstrations provides a basis for the LLM  to generate high-quality design points while leveraging infused broad design knowledge. Furthermore, the diversity of the common dataset contributed by BO enriches the contextual understanding of the LLM and allows it to more broadly search in the design space and prevent repetitive and redundant suggestions.

We evaluate the proposed framework on two different types of analog circuits and demonstrate notable improvements in design efficiency and effectiveness. By combining the exploratory capabilities of BO with the rich contextual knowledge of LLMs, our approach sets a new benchmark for automated analog circuit design, promising significant reductions in design time and improvements in outcome predictability.

\section{Background}

\subsection{Bayesian Optimization in Circuit Sizing}
Bayesian optimization (BO) \cite{bo-tutorial} is a powerful optimization algorithm to search the global optimum of a black-box function. For analog circuit sizing tasks, BO maximizes a figure of merit (FOM) function:
\begin{equation}
    x^*=\argmax_x\quad FOM(x)
\end{equation}
where $x \in \mathcal{X} \subset \mathrm{R}^d$ is a $d$-dimensional vector in a design space $\mathcal{X}$, $FOM:\mathrm{R}^d \to \mathrm{R}$ is a single value function that balances all circuit performance metrics, such as gain, unit gain frequency, etc.

BO leverages a probabilistic surrogate model to provide uncertainty quantification for the design space, and a corresponding acquisition function to trade off exploitation and exploration. In each iteration, BO determines the next query parameter set that maximizes the acquisition function, and then updates the surrogate model with the new simulation. Typically the surrogate is a Gaussian process (GP) \cite{gaussian-process}. The circuit design flow of BO is summarized in \cref{alg:bo}.

\begin{algorithm}
\caption{Bayesian Optimization for Analog Circuit Design}
\begin{algorithmic}[1]
\State \textbf{Input:} Initial data size $N_{init}$, number of iterations $N_{iter}$
\State $\bm{x}_{init} \gets$ Randomly sample  $N_{init}$ points from the design space $\mathcal{X}$
\State $\bm{FOM}_{init} \gets \text{Simulate the circuit with parameters } \bm{x}_{init}$
\State Fit an initial surrogate model $\gp$ on $(\bm{x}_{init}, \bm{FOM}_{init})$

\For{$t = 1$ \textbf{to} $N_{iter}$}
    \State Select the next point for simulation via maximizing the acquisition function $\alpha$: $x_t \gets \arg \max_{x} \alpha(x, \gp)$
    \State $FOM_t \gets \text{Simulate the circuit with parameters } x_t$
    \State \text{Update the surrogate model $\gp$ with $(x_t, FOM_t)$}
\EndFor

\State \textbf{return} The best observed point and value $(x^*, FOM^*)$
\end{algorithmic}\label{alg:bo}
\end{algorithm}

\subsection{Practice of Large Language Model in Analog Circuit Design}
Trained by the next token generation, Large language models (LLMs) have emergent abilities in many downstream tasks \cite{emergent-abilities}. Recent research \cite{ladac} shows that LLMs have prior knowledge about analog circuits, and the text box below demonstrates that LLMs understand codes of circuit netlists.


It is promising to leverage LLMs' prior knowledge about analog circuits for transistor sizing. The authors of \cite{ladac} design the first LLM agent LADAC for analog parameter design. They integrate two techniques: 1) in-context learning \cite{in-context-learning} for extracting information from design examples and a local design knowledge library, and 2) chain-of-thought \cite{chainofthought} for decision-making. Based on GPT-4, LADAC successfully discovers good circuit parameter sets of 3 analog circuits satisfying design specifications.

\tcbset{colback=white, colframe=black, fonttitle=\bfseries}
\begin{tcolorbox}[title=LLM's Understanding with Circuit Definition]
\textbf{User}: Analyze the structure of the following netlist of a two-stage differential amplifier:
\\
...
\\

\begin{tabular}{llllllll}
xn1 &n1 &v1 &n2 &0 &nfet &l=l1 &w=w1\\
xn2 &o1 &v2 &n2 &0 &nfet &l=l1 &w=w1\\
xp1 &n1 &n1 &vdd! &vdd! &pfet &l=l2 &w=w2\\
xp2 &o1 &n1 &vdd! &vdd!  &pfet &l=l2 &w=w2\\
\end{tabular}
\\
\\
...
\\
\\
\textbf{ChatGPT}:
\\
...
\\

\textbf{Stage One:} The first stage is a differential amplifier with a current mirror as the load. Here's how it's structured:

\textit{\quad 1. Differential Pair}: The first stage consists of transistors xn1 and xn2, which are N-channel MOSFETs forming a differential pair...

\textit{\quad 2. Current Mirror Load}: xp1 and xp2 are P-channel MOSFETs forming a current mirror load....

...
\end{tcolorbox}

\subsection{BO and LLM for Circuit Sizing: Pros and Cons}
BO has been extensively utilized for its ability to efficiently navigate complex design spaces, employing probabilistic GP surrogate models to balance exploration and exploitation. This balance is achieved through sophisticated acquisition functions, which prioritize areas of uncertainty that might yield high returns. Despite these strengths, BO's application in circuit sizing often faces challenges due to its generic approach, which lacks the incorporation of domain-specific knowledge. This deficiency can lead to a prevalence of non-viable solutions, i.e., parameters failed for success simulation. This phenomenon exacerbates in larger or more complex design landscapes, where the absence of tailored guidance becomes markedly detrimental.

Concurrently, LLMs have started to play a pivotal role in EDA by leveraging their vast reservoirs of encoded prior knowledge. These models apply advanced language understanding and generation capabilities to mimic expert-level decision-making processes. Through the use of few-shot learning and in-context learning techniques \cite{in-context-learning}, LLMs can quickly generate initial design points that are both innovative and feasible. However, their dependency on the quality and scope of the training data, as well as the few shot demonstration, often restricts the models to optimizing within a narrow region around provided examples, leading to potential sub-optimalities and poor exploration of broader design spaces.

To summarize, while BO excels in systematic exploration across a wide range of potential solutions, it lacks the intuitive, knowledge-driven insights provided by LLMs. Conversely, LLMs offer rapid generation of viable design points based on learned data and expert patterns but are constrained by their limited ability to generalize beyond familiar scenarios. This motivates us to propose a novel cooperative approach that synergies the predictive power of BO with the knowledge-driven capabilities of LLMs, aiming to harness the strengths of each while mitigating their individual limitations, as detailed in the next section.


\section{Methods}

\subsection{Overview of the ADO-LLM Framework}
The ADO-LLM framework integrates Large Language Models (LLMs) with Bayesian Optimization (BO) to address the complex multi-objective optimization challenges inherent in analog circuit sizing. This framework consists of four primary components: a Gaussian Process-based Bayesian Optimization (GP-BO) proposer, an LLM agent, a high-quality data sampler, and an HSPICE simulator, which evaluates the proposed design points.

Initially, the LLM agent is configured with circuit definition files and design specifications (see Figure \ref{fig:overview} top-left). Then it utilizes a zero-shot approach to leverage its relevant analog design knowledge obtained from the vast amount of model pre-training data for generating initial design points (see Figure \ref{fig:overview} bottom-left). These points, enriched with embedded domain insights, initiate an iterative loop where design points are proposed, evaluated via the simulator, and used to update the dataset, thereby enhancing the learning context for subsequent iterations. The overall architecture and optimization flow is illustrated in Figure \ref{fig:overview}.

\subsection{The GP-BO Proposer}
The Gaussian Process-based Bayesian Optimization (GP-BO) proposer is an integral part of the ADO-LLM framework, ensuring the diversity of newly generated design points. It employs a Gaussian Process (GP) with a Radial Basis Function (RBF) kernel as its surrogate model, fitted on the entire collected dataset to predict the Figure of Merit (FOM) across the extensive design space.

An acquisition function that maximizes expected improvement guides the GP-BO's search strategy, which balances between exploring new potential areas and exploiting regions with promising performance, following standard optimization practices \cite{mace}. This uncertainty-guided search ensures the diversity of the collected data and prevents the optimization process from becoming stuck at local optima.

\subsection{The LLM Agent}
The LLM agent is pivotal within the ADO-LLM framework and serves two key roles. Initially, it uses extensive pre-trained knowledge during the zero-shot initialization phase to suggest viable design points. In each subsequent optimization iteration, the agent utilizes a selectively chosen set of few-shot demonstrations—identified for their high Figure of Merit (FOM) from the diversely collected dataset—as well as domain-specific knowledge, to generate innovative design solutions.

The effectiveness of the LLM agent hinges on three critical components: the inherent domain expertise of the selected model, the quality of in-context learning designed to incorporate human expertise, and the quality of the demonstration data. The rationale behind the design choices for each of these components is further elaborated below:

\subsubsection{Model Choice:}
Currently, there is no suitable LLM specifically tailored for analog circuit design. Adapting or pretraining an LLM for this specialized field requires substantial data, and additional fine-tuning would be necessary to refine its ability to follow complex instructions. Given these challenges, we selected ChatGPT-3.5 Turbo as the backbone of the LLM agent, following the successful application of the ChatGPT API in other LLM-based EDA research \cite{gpt4aigchip}.

\subsubsection{In-context Learning:}
Circuit sizing is inherently a complex task that typically requires multiple steps of  reasoning. In ADO-LLM, we follow the chain of thoughts \cite{chainofthought} to prompt the LLM to decompose the problem into multiple steps:

\textbf{a. Interpreting the Circuit Definition:} Initially, the LLM is prompted to read and explain the given circuit definition, detailing the role of each transistor. Since ChatGPT-3.5 lacks specific domain knowledge, we incorporate human annotations in the prompt to minimize misinterpretations and potential error accumulation.

\textbf{b. Balancing Trade-offs in the Design Specifications:}
We prompt the LLM to describe how to meet \emph{all} design specifications for the interpreted circuit and explain the trade-offs between each objective. This step helps articulate the design knowledge pre-embedded within the large language dataset used for training of LLMs.

\textbf{c. Providing Few-shot Examples with Diverse Simulation Results: } Few-shot examples are provided to guide the LLM toward generating relevant and focused responses. In ADO-LLM, these examples include not only design points and simulated metrics but also higher-level feedback from simulators, such as transistor operational regions. This approach enables the LLM to base decisions on comprehensive data rather than just numerical outputs.

\textbf{d. Injecting Human Expertise with Design Principles:} We also utilize additional instructions termed ``design principles'' to improve the generation quality of the LLM agent. The adopted design principles instruct the LLM to ensure that all transistors operate within specific desired regions when proposing design points. These principles help the LLM utilize high-level simulator feedback effectively, while also preventing it from merely regurgitating trivial solutions from its training data.

Finally, the LLM is prompted to generate a new design point following the thought of the previous steps with the given format and parameter range constraints. We adopt a parser that ensures the correct formatting of the response and requests the LLM to regenerate if the requirements are not satisfied.

\subsubsection{High-Quality Data Sampler:} The diversity of design points has been enhanced by the GP-BO proposer. Within ADO-LLM, however, we selectively use only a subset of the full dataset as few-shot examples for the LLM agent. This selection is facilitated by a high-quality data sampler that identifies and samples the top-performing demonstrations based on the top-k Figure of Merit (FOM). This method ensures the chosen examples are not only diverse—thereby preventing the LLM from converging on local optima—but also of high value, providing potent models for the LLM to emulate. Consequently, this strategic sampling empowers the LLM to generate relevant and innovative responses, leveraging the most effective designs to propose new design points for future evaluation.


\section{Experiments}
\begin{table}[H]
\centering
\caption{Hyper-parameters of the proposed \ourmethod}
\begin{tabular}{l l l}
\toprule
\textbf{Sub-model} &
\textbf{Hyper-parameter}            & \textbf{Value}                         \\ \midrule
LLM Agent & \# Queries per Step          &   1                      \\
    & \# Top Examples in Context  &   5                   \\
    & LLM Version                   & GPT 3.5                        \\ 
    & Temperature                   & 0.5                         \\
    & Context Window Length         & 16k \\
    & Max Token Generation Length  & 1,000                    \\
   \midrule
GP-BO  &     \# Queries per Step      & 4                 \\
    &     Kernel Function        & RBF                      \\
    &     Acquisition   & qEI \cite{qEI}                        \\
    &     Acquisition Optimizer     & L-BFGS                   \\
\bottomrule
\label{tab:ado-llm-config}
\end{tabular}
\end{table}

We benchmark ADO-LLM on the sizing of two distinct and fundamental analog circuits: (1) a two-stage differential amplifier and (2) a hysteresis comparator. Initially, we demonstrate that the cooperative interaction between the LLM and Bayesian Optimization (BO) significantly enhances search efficiency compared to using either method alone. Subsequently, we conduct several ablation studies to analyze the effectiveness of different components within the system.


\paragraph{\ourmethod\ Model Settings}
\ourmethod\ has two parallel optimization models: the GP-BO  model and the LLM agent. The configurations of both models are presented in \cref{tab:ado-llm-config}. \ourmethod\ starts with 5 initial examples predicted by the LLM agent. In each data query iteration, we let the LLM agent propose 1 candidate, based on the designed prompt and top 5 data points; and let GP-BO provide 4 candidates from a qEI acquisition function \cite{qEI}, constructed over the posterior prediction of a GP surrogate. 

\paragraph{Baseline Models} 
We compare \ourmethod\ with two baselines: the GP-BO model with random initialization and a single LLM proposer that uses zero-shot initialization to generate starting points. The number of initial design data is 5 for both methods. To ensure a fair comparison, the total number of evaluations of each method is set to be the same as that of \ourmethod.

\begin{figure}
    \centering
    \includegraphics[width=0.47\textwidth]{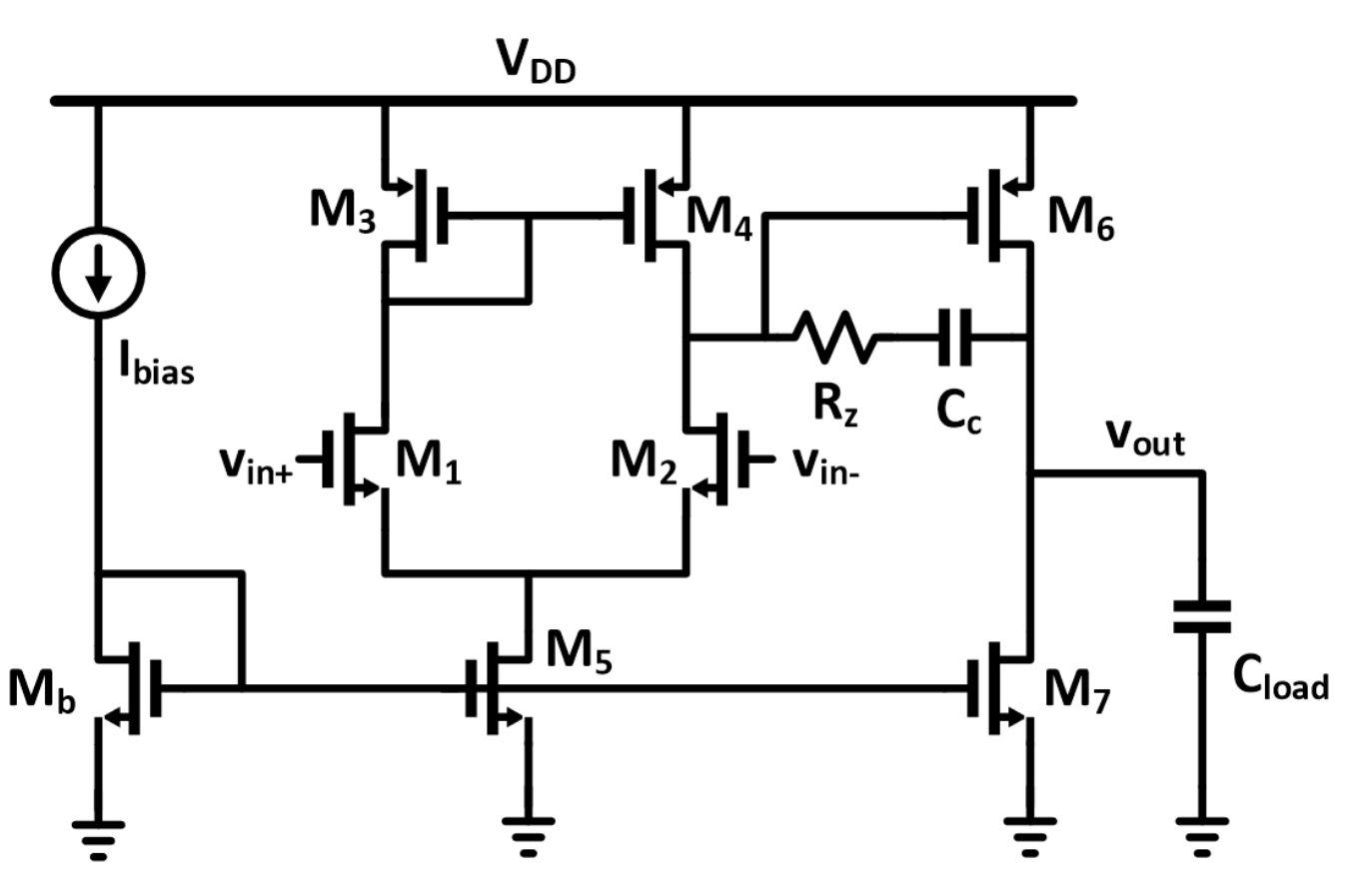}
    \caption{The circuit schematic of the two-stage differential amplifier}
    \label{fig:ckt-amp2}
\end{figure}

\begin{table}[]
\centering
\caption{Design configuration of a two-stage differential amplifier under a commercial 90nm CMOS technology}
\begin{tabular}{l l}
\toprule
\textbf{Attribute}            & \textbf{Value}                         \\ \midrule
Transistor Width Range                   & [120nm, 50$\mu$m]                        \\ 
Transistor Length Range                  & [80nm, 1$\mu$m]                          \\ 
Resistance Range              & [10$\Omega$, 100k$\Omega$]                        \\ 
Capacitance Range             & [10fF, 100pF]                    \\ \midrule
Gain-Bandwidth Product (Spec)  & $\geq$ 1MHz                                 \\ 
Gain (Spec)                    & $\geq$ 60dB                                   \\ 
Common-Mode Rejection Ratio (Spec) & $\geq$ 75dB                              \\ 
Phase Margin (Spec)            &$\geq$ 60\degree                                   \\ 
Power Consumption (Spec)       & $\leq$ 30$\mu$W                              \\ 
\midrule
Gain-Bandwidth Product (Norm)  & [0MHz, 10MHz]                               \\ 
Gain (Norm)                    & [0dB, 60dB]                                \\ 
Common-Mode Rejection Ratio (Norm) & [0dB, 80dB]                          \\ 
Phase Margin (Norm)            & [0\degree, 45\degree]                               \\ 
Power Consumption (Norm)       & [0$\mu$W, 30$\mu$W]                            \\ 
\midrule
Gain-Bandwidth Product (Failed)  & -10MHz                                  \\ 
Gain (Failed)                    & -60dB                                     \\ 
Common-Mode Rejection Ratio (Failed) & -80dB                                \\ 
Phase Margin (Failed)            &-180\degree                                     \\ 
Power Consumption (Failed)       & 80$\mu$W                                \\ 
\bottomrule
\label{tab:ckt-amp2-info}
\end{tabular}
\end{table}

\subsection{Two-Stage Differential Amplifier} \label{sec:rslt-amp2}
\paragraph{Experimental Settings}
 The two-stage differential amplifier is depicted in \cref{fig:ckt-amp2}. It has 18 parameters: the length and the width of 8 transistors ($M_1$ to $M_7$, and $M_b$), the resistance of a feedback resistor ($R_z$), and the capacitance of a compensation capacitor ($C_c$). The number of free parameters is 14, where some transistors share the same size ($M_1$ and $M_2$;$M_3$ and $M_4$). The design space of circuit parameters is illustrated in \cref{tab:ckt-amp2-info}. We use Synopsys HSPICE to simulate the performance of designed circuits, under a commercial 90 nm CMOS technology.

We aim to design a circuit that satisfies the specifications of several performance metrics, including gain-bandwidth product (GBW), gain, common-mode rejection ratio (CMRR), phase margin (PM), and power consumption. These specifications are listed in \cref{tab:ckt-amp2-info}. To facilitate GP-BO for this multi-objective optimization task, we construct FOM as a linear combination of the considered metrics:
\begin{table*}[]
    \centering
\caption{The performance of the best parameter set found by each method for the two-stage differential amplifier}
\begin{tabular}{l c c c c c c c c c}
\toprule
\makecell{Method} &  \makecell{\# Simulation \\ Init. + Batch $\times$ Iter} & \makecell{Gain (dB) \\ $\geq 60$} & \makecell{CMRR (dB) \\ $\geq 75$} & \makecell{GBW (MHz) \\ $\geq 1$} & \makecell{PM ($\degree$) \\ $\geq 60$} & \makecell{Power ($\mu$W) \\ $\leq 30$} & \makecell{FOM} & \# Missed Spec. \\ 
\midrule
GP-BO        & $5+5\times 20$     &  {27.12 (\xmark)}     &  \textbf{106.08}     &  4.63     &   \textbf{97.76}    &    27.02   &  1.89   & 1            \\ 
GP-BO        & $5+5\times 80$     &  {\textbf{63.09}}     &  79.57     &  7.22     &   94.28    &    65.76 (\xmark)  &  2.10    & 1         \\ 
LLM agent   & $5+1\times 100$    &  60.35     &  73.04     &   \textbf{93.09}    &   87.02    &   {39.00 (\xmark)}    & 0.20      & 1          \\ 
\midrule
\ourmethod   & $5+5\times20$   &  60.83     &  78.38     &   1.35    &  92.29     &   \textbf{19.79}    &  \textbf{3.52}       & \textbf{0}        \\ 
\bottomrule
\end{tabular}    
    \label{tab:amp2-main-rslt}
\end{table*}

\begin{equation}
    FOM =   \widetilde{GBW}^{b} + \widetilde{Gain}^{b} + \widetilde{CMRR}^{b} + \widetilde{PM}^{b} -\widetilde{Power} 
\end{equation}
 The operator $\widetilde{ \cdot }$ is a function to check whether the specification is satisfied, and then to normalize the metric. It is defined as
\begin{equation}
    \widetilde{m} :=
    \begin{cases}
        \frac{m-m_\text{min}}{m_\text{max}-m_\text{min}} & \text{if } m \text{ hits } m_\text{spec} \\
        \frac{m_\text{failed}-m_\text{min}}{m_\text{max}-m_\text{min}} & \text{otherwise}
    \end{cases}
\end{equation}
for a metric $m$ where the normalization range and the failed value of each metric are listed in \cref{tab:ckt-amp2-info}.

The superscript ${b}$ means the bounded value, i.e., a metric does not need to exceed far beyond its design specification. This bound is set to 2 for the normalized values of GBW, Gain, CMRR, and PM.

\paragraph{Results} For each method, we present its performance metrics, FOM, and number of missed specifications in \cref{tab:amp2-main-rslt}. The best metric across all methods is bold, and a cross mark is appended to a metric when it fails to meet the corresponding design specification. Compared with GP-BO and LLM agent, \ourmethod\ achieves the largest FOM, which is even better than that acquired by GP-BO after 4x iterations of simulation. More importantly, only \ourmethod\ satisfies all design specifications. While the baselines GO-BO and the LLM agent excel in optimizing several circuit performance metrics, they fail to balance these metrics with a limited simulation budget. This result demonstrates the data efficiency of \ourmethod\ in analog circuit design tasks.

\subsection{Hysteresis Comparator} \label{sec:rslt-comp}
\paragraph{Experimential Settings} We aim to optimize a hysteresis Comparator in \cref{fig:ckt-comp} via adjusting the size of 12 transistors ($M_1$ to $M_{11}$ and $M_b$). It has 12 free parameters to be determined, including 6 values of length and width of transistors. We use Synopsys HSPICE to simulate the performance of designed circuits, under a commercial 90 nm CMOS technology.

The metrics to be optimized are gain, unit gain frequency (UGF), Hysteresis Error ($\text{V}_{\text{hys\_err}}$), voltage offset ($\text{V}_{\text{offset}}$), and power.  We construct $FOM$ as follows:
\begin{figure}
    \centering
    \includegraphics[width=0.45\textwidth]{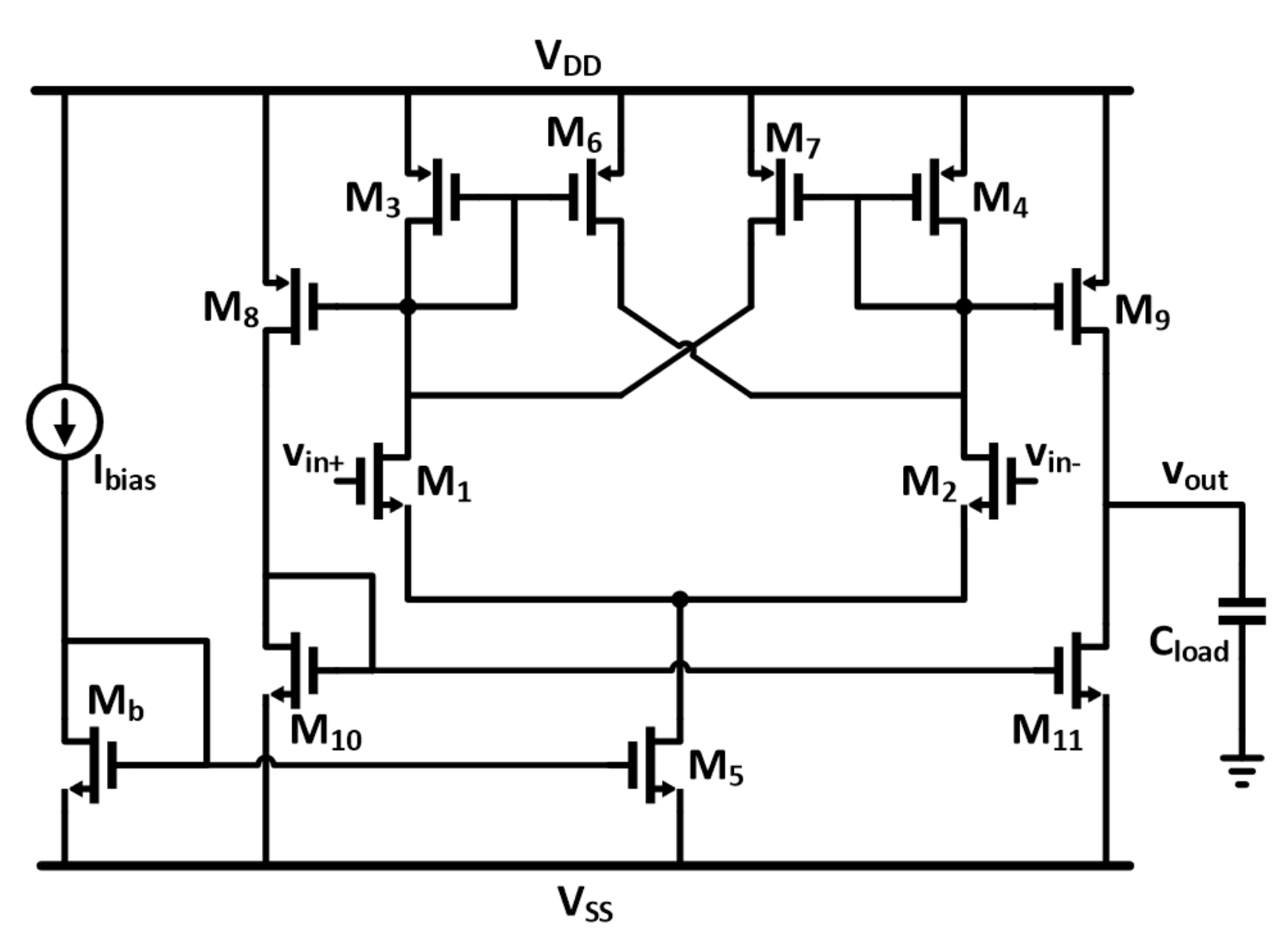}
    \caption{The circuit schematic of a hysteresis comparator}
    \label{fig:ckt-comp}
\end{figure}
\begin{table}[]
\centering
\caption{Design configuration of a hysteresis comparator under a commercial 90nm CMOS technology}
\begin{tabular}{l l}
\toprule
\textbf{Attribute}            & \textbf{Value}                         \\ \midrule
Transistor Width Range                   & [90nm, 200$\mu$m]                        \\ 
Transistor Length Range                  & [90nm, 1$\mu$m]                          \\ 
\midrule
Gain (Spec)                    & $\geq$ 60dB                                   \\ 
Unit Gain Frequency (Spec)  & $\geq$ 10MHz                                 \\ 
Absolute Hysteresis Error (Spec)            &$\leq$ 300mV                                  \\ 
Absolute Voltage Offset (Spec) & $\leq$ 20mV                              \\ 
Power Consumption (Spec)       & $\leq$ 150$\mu$W                              \\ 
\midrule
Gain (Norm)                    & [0dB, 60dB]                                \\ 
Unit Gain Frequency (Norm)  & [0mHz, 10MHz]                               \\ 
Hysteresis Error (Norm)            & [0mV, 300mV]                               \\ 
Absolute Voltage Offset (Norm) & [0mV, 20mV]                          \\ 
Power Consumption (Norm)       & [0$\mu$W, 150$\mu$W]                            \\ 
\midrule
Gain (Failed)                    & -60dB                                     \\ 
Unit Gain Frequency (Failed)  & -10MHz                                  \\ 
Hysteresis Error (Failed)            & 600mV                                     \\ 
Absolute Voltage Offset (Failed) & 40mV                                \\ 
Power Consumption (Failed)       & 300$\mu$W                                \\ 
\bottomrule
\label{tab:ckt-comp-info}
\end{tabular}
\end{table}

\begin{table*}[]
    \centering
\caption{The performance of the best parameter set found by each method for a hysteresis comparator}
\begin{tabular}{l c c c c c c c c}
\toprule
\makecell{Method} &  \makecell{\# Simulation \\ Init. + Batch $\times$ Iter} & \makecell{Gain (dB) \\ $\geq 60$} & \makecell{UGF (MHz) \\ $\geq 10$} & \makecell{$\text{V}_{\text{hys\_err}}$ (mV) \\ $  \leq 300$} & \makecell{$\text{V}_{\text{offset}}$ (mV) \\ $ |{\cdot}| \leq 20$} & \makecell{Power ($\mu$W) \\ $\leq 150$} & \makecell{FOM} & \# Missed Spec. \\ 
\midrule
GP-BO        & $5+5\times20$     &  {55.47 (\xmark)}     &  8.42 (\xmark)     &  199.42      &   -3.95     &    \textbf{77.70}   &  -3.38   & 2            \\ 
GP-BO        & $5+5\times80$     &  {30.18 (\xmark)}     &  10.10     &  186.33     &   3.55    &    94.31  &  -1.42 & 1            \\ 
LLM agent   & $5+1\times100$    &  40.52 (\xmark)     &  \textbf{13.66}     &   161.14    &   7.36    &   121.37    &  -1.35      & 1         \\ 
\midrule
\ourmethod   & $5+5\times20$   &  \textbf{60.83}     &  12.04     &   \textbf{159.83}    &  \textbf{-1.00}     &   109.89    &  \textbf{0.90}      & \textbf{0}         \\ 
\bottomrule
\end{tabular}    
    \label{tab:comp-main-rslt}
\end{table*}

\begin{equation}
    FOM =   \widetilde{Gain}^{b} + \widetilde{UGF}^{b} - \widetilde{V_{hys\_err}} - \widetilde{|V_{offset}|} -\widetilde{Power} 
\end{equation}
 \cref{tab:ckt-comp-info} presents our design specifications and configurations. The upper bound is set to 2 for the normalized values of Gain and UGF in our experiments.

\paragraph{Results} \cref{tab:comp-main-rslt} shows the performance metrics, the FOM, and the number of missed design specifications of each optimization approach. The result of \ourmethod\ in the comparator is consistent with that in the amplifier: it achieves the highest FOM, and only our method fulfills the specification demands.

\begin{table}[]
    \centering
    \caption{The best design of GP-BO with different initializations}
    \begin{tabular}{lccc}
    \toprule
         Method & Circuit & FOM & \# Missed Spec. \\
     \midrule
         GP-BO with random init. & Amp2  & 1.89 & 1 \\
         GP-BO with LLM's init. & Amp2 &  \textbf{2.20} & 1 \\    
    \midrule
         GP-BO with random init. & Comp & -3.38 & 2 \\
         GP-BO with LLM's init. & Comp &  \textbf{-1.64} & \textbf{1} \\    
    \bottomrule
    \end{tabular}
    \label{tab:ablation-zeroshot}
\end{table}

\subsection{The significance of Zero-Shot Initialization} \label{sec:rslt-zeroshot}

We conduct an ablation study to demonstrate that the zero-shot initialization with the LLM proposer plays a vital role in generating a good starting point and hence accelerating the optimization process, compared with random initialization.

\paragraph{Experimental Settings} 
We employed Gaussian Process-based Bayesian Optimization (GP-BO) for the design of the specified amplifier and comparator circuits. We compared two initialization strategies: uniform sampling from the entire search space and zero-shot initialization provided by the LLM proposer. We configured the experiments with 5 initial points, a batch size of 5 queries, and a total of 20 iterations.

\paragraph{Results}
\cref{tab:ablation-zeroshot} shows the best FOM acquired by GP-BO, where the abbreviation Amp2 stands for the two-stage amplifier, and Comp stands for hysteresis comparator. The initial data from the LLM contributes to a better FOM in either circuit, which is even comparable with that acquired after 80 iterations, as listed in \cref{tab:amp2-main-rslt} and \cref{tab:comp-main-rslt}. This result demonstrates that the domain-specific prior knowledge of the LLM is able to provide good initialization parameter sets in potential high-value regions for analog circuit design tasks.

\subsection{The Effectiveness of In-Context Learning and Proposed Sampler}\label{sec:rslt-in-context-learning}

The quality of few-shot demonstrations during in-context learning plays a critical role in eliciting high-quality responses from the LLM agent. We conducted experiments to validate the necessity and impact of these demonstrations.
\begin{table}[]
    \centering
    \caption{The best design of the LLM agent with different in-context examples}
    \begin{tabular}{lcccc}
    \toprule
         Method & Circuit & Sampler & FOM & \makecell[c]{\# Missed \\ Spec.} \\
     \midrule
         LLM agent w/o ICL & Amp2 & NA& -1.78  & 3 \\
         LLM agent with ICL & Amp2 & Rand 5 & -0.63 & 2 \\
         LLM agent with ICL & Amp2 & Top 5 & \textbf{0.20} & \textbf{1} \\
    \midrule
         LLM agent w/o ICL & Comp & NA & -2.29 &  2 \\
         LLM agent with ICL & Comp & Rand 5 & -2.53 & 3 \\
         LLM agent with ICL & Comp & Top 5 & \textbf{-1.35} & \textbf{1} \\
    \bottomrule
    \end{tabular}
    \label{tab:ablation-in-context-learning}
\end{table}

\paragraph{Experimental Settings}
We design two variants of the LLM agent: the first LLM agent is not provided with any demonstration.  The second agent learns from few-shot demonstrations that are uniformly sampled from the collected dataset. We compare these two variants with the proposed LLM agent which is provided with few-shot demonstrations with the highest top-k FOM. For all three variants, the number of sampled demonstrations if there is any in each iteration is 5, and the total simulation evaluation budget is 105.

\paragraph{Results} \cref{tab:ablation-in-context-learning} reports the best design example of the LLM agent variants in optimizing the amplifier and the comparator. Here the abbreviation ICL stands for in-context learning. While using random sampling to provide demonstrations improves  the performance of the amplifier circuit, this strategy degrades the zero-shot capacity of the LLM agent when optimizing the comparator. With the high-quality design demonstrations provided by the proposed sampler, the LLM agent achieves superior performance in both the  FOM and the number of specifications met, demonstrating the effectiveness of the in-context learning and our sampling strategy for analog circuit design.

\section{Related Work}
\subsection{ML-based Automated Analog Circuit Design} 

Machine learning has been a popular tool for analog circuit design for decades \cite{ml4ckt-design}. Past ML approaches primarily utilize either reinforcement learning (RL) \cite{rl-survey} or Bayesian optimization (BO) \cite{bo-tutorial, bo4ckt-design}.

RL-based algorithms focus on training a design agent that outputs circuit parameters aimed at maximizing the ultimate reward. For instance, GCN-RL \cite{gcn-rl} employs a graph convolutional neural network (GCN) \cite{gcn} to interpret circuit topology information, while Prioritized RL \cite{prioritized-rl} introduces a non-uniform sampler to prioritize design parameters from potentially high-value areas.

On the other hand, BO-based methods balance the exploration and exploitation of the design space by optimizing an acquisition function over the Gaussian process \cite{gaussian-process} posterior predictions. Techniques like MACE \cite{mace} and pHCBO \cite{phcbo} use ensembles of multiple acquisition functions. MACE selects query batches from the Pareto front computed by DEMO \cite{demo}, and pHCBO constructs batches from the optimal points of each acquisition function.

However, both BO and RL approaches face challenges in meeting multi-objective design specifications because they typically reduce the analog circuit design task to a single-objective global optimization problem, focused on maximizing a Figure of Merit (FOM), a weighted linear combination of performance metrics. This reduction creates a dilemma: a simplistic FOM cannot adequately cover all design specifications, yet a complex FOM becomes too cumbersome to optimize effectively. Thus, the design of the FOM critically influences the overall performance of circuit sizing.

\subsection{Large Language Models for Electronic Design Automation} 
The integration of large language models (LLMs) into the field of electronic design automation (EDA) has been a subject of considerable interest, exploring various methodologies and applications. Broadly, the utilization of LLMs in EDA can be classified into two distinct approaches: the use of externally hosted APIs and the development of domain-adapted LLMs specifically tailored for EDA tasks.

In terms of application diversity, significant contributions have been made across several key areas. First, the completion of Verilog code, where LLMs assist in generating and suggesting code segments to streamline the hardware design process, shows promise in enhancing productivity and reducing error rates \cite{verilogeval, llm_verilog_benchmark}. Second, in the design of accelerators, LLMs have been employed to optimize the architectural decisions, potentially leading to more efficient processing for specific tasks \cite{gpt4aigchip, chipchat}. Another noteworthy application is in the completion of EDA software code, which not only aids in software development but also ensures that tools are more adaptable to the needs of hardware engineers \cite{chateda}.

Further, the use of LLMs in debugging represents a significant shift towards more intelligent troubleshooting methods in circuit design, enabling quicker identification and rectification of errors in both hardware and software components \cite{chipnemo}. Lastly, in the domain of analog circuit sizing, LLMs have demonstrated their utility by automating the adjustment of component sizes for optimal performance, thus facilitating more efficient design cycles \cite{ladac}.

\section{Limitation and Future Works}
In this paper, we proposed ADO-LLM, which integrates Bayesian Optimization (BO) with Large Language Models (LLMs) to enhance the efficiency of analog circuit sizing and has demonstrated promising results on two different analog circuit sizing problems. However, like all methodologies, it is not without limitations, which open avenues for further research and refinement.

\textbf{1. Dependency on External LLM APIs:} Currently, our system utilizes LLMs accessed via closed-source ChatGPT APIs, which can be costly and lack transparency. Furthermore, these models are not specifically tailored to the nuances of circuit design, potentially limiting their effectiveness and applicability. Future work could explore the development and integration of domain-adapted LLMs\cite{chipnemo} that are specifically trained on electronic design automation tasks. Additionally, incorporating techniques like Retrieval-Augmented Generation (RAG) could enable the system to leverage more precise, domain-specific knowledge, enhancing the accuracy and relevance of the solutions generated.

\textbf{2. Optimality and Scalability of GP in BO:} The Gaussian Processes (GP) used in our current Bayesian Optimization (BO) setup effectively manage the design spaces of the circuits examined in this study. However, they may not scale well for more complex designs. In practical applications, as design spaces expand, the computational overhead and the inefficacy of optimizing all parameters simultaneously become significant constraints. A promising direction for future research is to implement hierarchical optimization strategies, where only a subset of parameters is optimized in each iteration, mimicking the approach of human experts. This could potentially increase both the efficiency and reliability of the design process. We leave these interesting directions to be explored in the future.

\section{Acknowledgement}
This material is based upon work supported by the National Science Foundation under Grants No. 1956313 and No. 2334380.



\bibliographystyle{ACM-Reference-Format}
\bibliography{sample-base}


\end{document}